\documentclass{article}

\usepackage[final,nonatbib]{nips_2017}

\usepackage[utf8]{inputenc} 
\usepackage[T1]{fontenc}    
\usepackage{url}            
\usepackage{booktabs}       
\usepackage{amsfonts}       
\usepackage{nicefrac}       
\usepackage{microtype}      
\usepackage{comment}

\usepackage{times}
\usepackage{epsfig}
\usepackage{graphicx}
\usepackage{amsmath}
\usepackage{amssymb}

\usepackage{hyperref}

\begin{document}

\title{MemexQA: Visual Memex Question Answering}

\author{Lu Jiang$^1$\thanks{Currently at Google.}, Junwei Liang$^1$, Liangliang Cao$^2$, Yannis Kalantidis$^3$\thanks{Currently at Facebook.}\\
\textbf{Sachin Farfade$^3$, Alexander Hauptmann$^1$}\\
$^1$Carnegie Mellon University, $^2$Customer Service AI, $^3$Yahoo Research
}

\maketitle

\newcommand{\yannis}[1]{{\color{cyan}{(YANNIS\@: #1)}}}
\newcommand{\lu}[1]{{\color{red}{(Lu\@: #1)}}}
\newcommand{\llc}[1]{{\color{green}{(Liangliang\@: #1)}}}
\newcommand{\sachin}[1]{{\color{blue}{(Sachin\@: #1)}}}
\newcommand{\alert}[1]{{\color{red}{#1}}}

\begin{abstract}

This paper proposes a new task, MemexQA: given a collection of photos or videos from a user, the goal is to automatically answer questions that help users recover their memory about events captured in the collection. Towards solving the task, we 1) present the MemexQA dataset, a large, realistic multimodal dataset consisting of real personal photos and crowd-sourced questions/answers, 2) propose MemexNet, a unified, end-to-end trainable network architecture for image, text and video question answering. Experimental results on the MemexQA dataset demonstrate that MemexNet outperforms strong baselines and yields the state-of-the-art on this novel and challenging task. The promising results on TextQA and VideoQA suggest MemexNet's efficacy and scalability across various QA tasks.
\end{abstract}


\section{Introduction}
\label{sec:intro}

A typical smartphone user may take hundreds of photos when on vacation or attending an event. Within only a few years, one ends up accumulating dozens of thousands of photos and many hours of video, that capture cherishable moments from one's past. 
A recent study~\cite{jiang17delving} hints that people utilize personal photos as a mean of recovering pieces from their memories. 
In this paper, we propose \textit{MemexQA}\footnote{\textit{Memex} was first posited by Bush in 1945~\cite{bush1945atlantic} as an enlarged intimate supplement to an individual's memory.}, a new AI task for question answering: given a collection of photos or videos from a user, the goal is to automatically answer questions that help recover his/her memory about the event captured in the collection. With conversational bots like Siri and Alexa becoming ubiquitous, natural language questions seem to be the best way of going down memory lane. A system utilizing MemexQA can be regarded as a personal assistant of one's past memories and can answer questions like ``When did we last go hiking/camping?'' or `` Who had a birthday party in January 2017?'', as long as the answer clues are captured by the user's photos or videos.

In order to derive an answer for such questions, we rely not only on visual understanding of single photos or videos, but the photo collection as a whole and select the corresponding photos and videos that may support the answers. This extends the scope of classical Visual QA (VQA)~\cite{antol2015vqa} task, which is designed to answer natural language questions for an image, questions that a toddler or an alien would have trouble answering. MemexQA also answers questions based on images. The questions however may possibly require reasoning over multiple images. The answers for MemexQA are not trivial even for an adult to answer manually, as it would require going through a large amount of media first. In  our experiments, human participants spend at least \textit{10x longer} time  to answer a MemexQA question than answering a VQA question.  Therefore, we hypothesize that MemexQA models can greatly help users recall their past memories. However, MemexQA is more challenging than classical VQA in the following aspects:

\vspace{-7pt}
\begin{description}
\setlength\itemsep{-2pt}
\item [Collective reasoning] MemexQA adds a localization component across media documents. One needs to inspect multiple items to find clues for the media that support the answer to a given question, while also properly understanding the (visual or other) content of each document.
\item [Cross-modal reasoning] MemexQA utilizes multimodal knowledge to provide an answer, which may come from the visual content, incomplete titles or tags, timestamps or GPS. It therefore requires some form of reasoning across multimodal information from a large scale dataset. Of course this can be extended to any extra annotations that a collection might have, \emph{e.g.} face/person identification, action recognition, sentiment analysis results.
\item [Dynamic answer space / memory] The answer space for MemexQA is somewhere between the closed set of most current VQA tasks and open-ended QA, as it is conditioned on the database. However, not only this space is not predefined and different among users, it also grows as the database increases. This is far larger than any current VQA task. Furthermore, MemexQA has an information retrieval component and takes into account the ever-growing dynamic memory of a user's media collection. This implicitly connects the task to personalization as the same question would give totally different answers for 2 different users.
\end{description}




\begin{figure}[t!]
\centering
\vspace{-3mm}
\includegraphics[width=0.88\linewidth, height = 58mm]{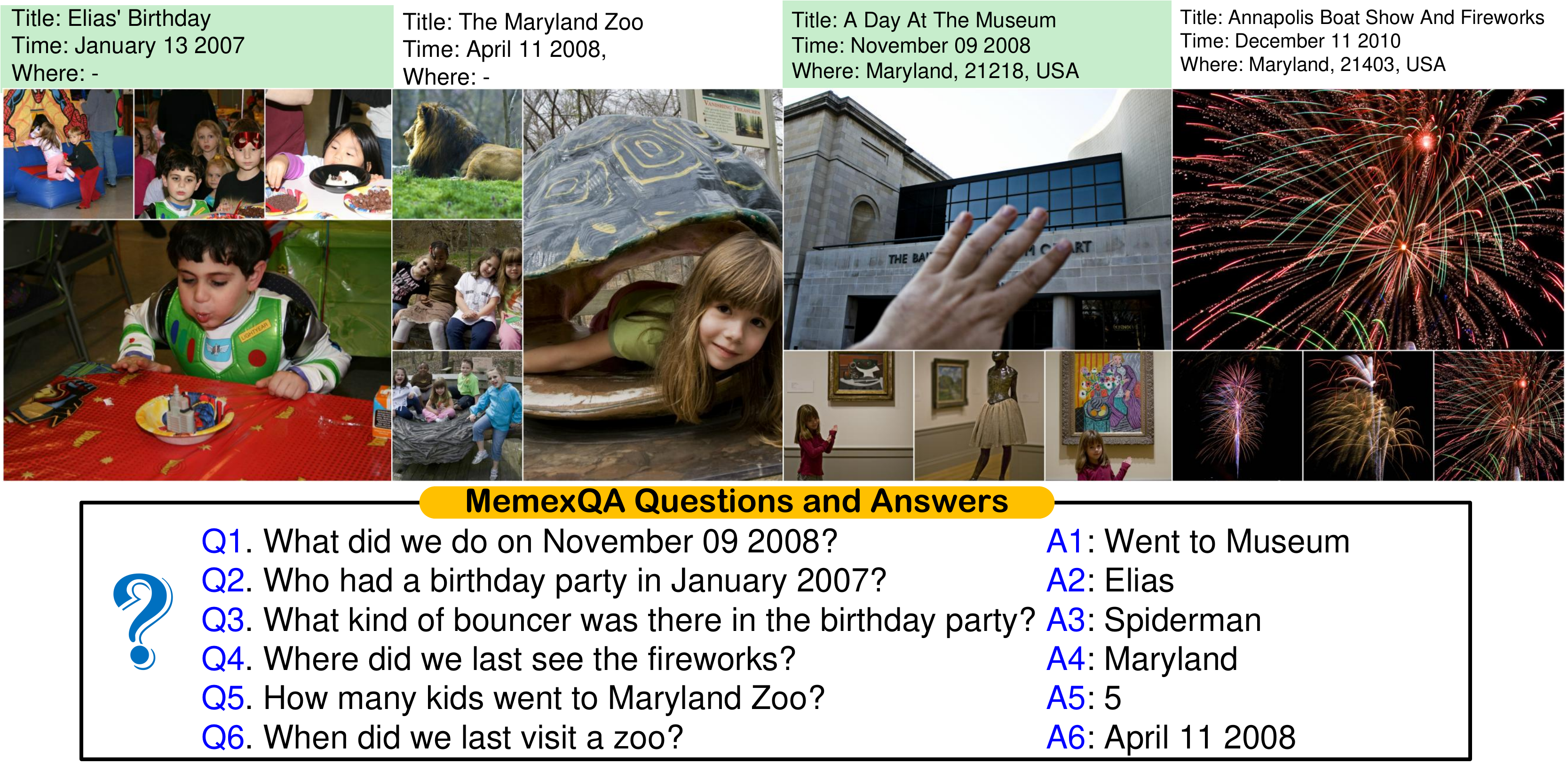}
\caption{\label{fig:intuition}MemexQA examples. Top: sampled personal photos of a Flickr user, each photo collage corresponds to an album. Bottom: representative questions and answer pairs. 
}
\vspace{-5mm}
\end{figure}

\vspace{-7pt}
In this paper we present a novel approach towards solving the MemexQA task. Given the absence of data to tackle this problem, we first crowdsourced questions and answers for personal Flickr images~\cite{huang2016visual}, organized in user albums. That results in the\textit{MemexQA dataset} (\url{https://memexqa.cs.cmu.edu/})
consisting of 13,591 personal photos from 630 albums and 101 Flickr users and 20,860 crowdsourced questions. Examples of questions, answers and photos from the dataset are illustrated in Fig.~\ref{fig:intuition}.


We further propose \textit{MemexNet}, an united network architecture for image, text and video question answering. The proposed MemexNet is an end-to-end trainable network which is learned together with off-the-shelf image/video engines and is of configurable modular networks for answer inference. To summarize, our contributions are the following: 1) We introduce a new AI task, MemexQA, as well as a large, croudsourced dataset that consists of question and answer pairs corresponding about not just a single image but also a collection of images. 2) We propose a novel approach to tackle MemexQA via a unified, end-to-end trainable model. Extensive experimental results on the MemexQA dataset and demonstrate that the proposed method outperforms all strong baselines and yields the state-of-the-art on this very challenging dataset. The promising results on TextQA and VideoQA suggest the efficacy and scalability of the proposed method across various QA tasks.

We consider MemexQA to be a multi-disciplinary AI task as it combines computer vision, natural language processing and information retrieval and extraction. We believe that this combination will enable us to scale  to real-world applications and ever-growing user media collections. Our study also provides a first but promising solution to answer questions based on large-scale video repository.


\vspace{-2mm}
\section{Related Work}
\label{sec:related}
\vspace{-2mm}


\noindent
\textbf{Visual QA}
VQA was first proposed by Antol \emph{et al.}
 in \cite{antol2015vqa}
and has received a large amount of attention in recent years \cite{yang2016stacked, xiong2016dynamic,LuXPS16, LuYBP16, FukuiPYRDR16, xu2015show, wu2016ask, WangWSHD16}.
The goal of VQA is to answer open-ended questions about a given image. The challenge of VQA lies in   understanding both the image semantics \cite{xu2015show} and the knowledge base \cite{wu2016ask, WangWSHD16}. The difference between MemexQA and VQA roots in the scalability, where MemexQA models must digest the whole collection of personal images and videos before answer the questions.


\noindent
\textbf{Text QA}
An important contributor to the recent advancement in Text QA has been the availability of various datasets. Our MemexQA is motivated by Watson \cite{ferrucci2010building} and Memory Networks~\cite{weston2014memory}, of which both learn a text based matching model for either open domain or close domain questions. Our MemexQA considers problem in a different field with multi-modalities including images, videos, and meta data. However, it can  be applied to Text QA problems too. The recently released SQuAD~\cite{rajpurkar2016squad} dataset with over 100,000 question-answer pairs has sparkled many studies on end-to-end machine comprehension neural models. 
Match-LSTM~\cite{wang2016machine} proposed a method to match the context with the question and then utilized a modified Pointer Network to find the answer positions.
DCN~\cite{xiong2016dynamicb} introduced a dynamic co-attention network to capture the interactions between the context and the question and used a dynamic decoder to predict the beginning and the ending points of the answer.
MPCM~\cite{wang2016multi} proposed a multi-perspective context matching model. The bi-directional attention flow (BiDAF)~\cite{seo2016bidirectional} used a bi-directional attention mechanism to get question-aware context representation, while a variant of Memory Network \cite{dhingra2016gated,shen2016reasonet} was used to calculate the attention vector between query and context. We compare some of these recent models in the Text QA task in Section~\ref{sec:exp_textqa}

\noindent
\textbf{Memex}
The term ``Memex'' was first posited by V. Bush in 1945 as an enlarged intimate supplement to an individual's memory. Bush envisioned the Memex as a device in which individuals would compress and store all of their information, mechanized so that it may be consulted with exceeding speed and flexibility~\cite{bush1945atlantic}. The concept of the Memex influenced the development of early hypertext systems, eventually leading to the creation of the World Wide Web~\cite{davies2011still} and the personal search system~\cite{gemmell2002mylifebits}. The proposed MemexQA advances the traditional Memex system to answer questions about the personal photo and video content.


\vspace{-2mm}
\section{The MemexQA dataset}
\vspace{-2mm}

\begin{figure}[t!]
\centering
\vspace{-3mm}
\includegraphics[width=0.99\linewidth, height = 60mm]{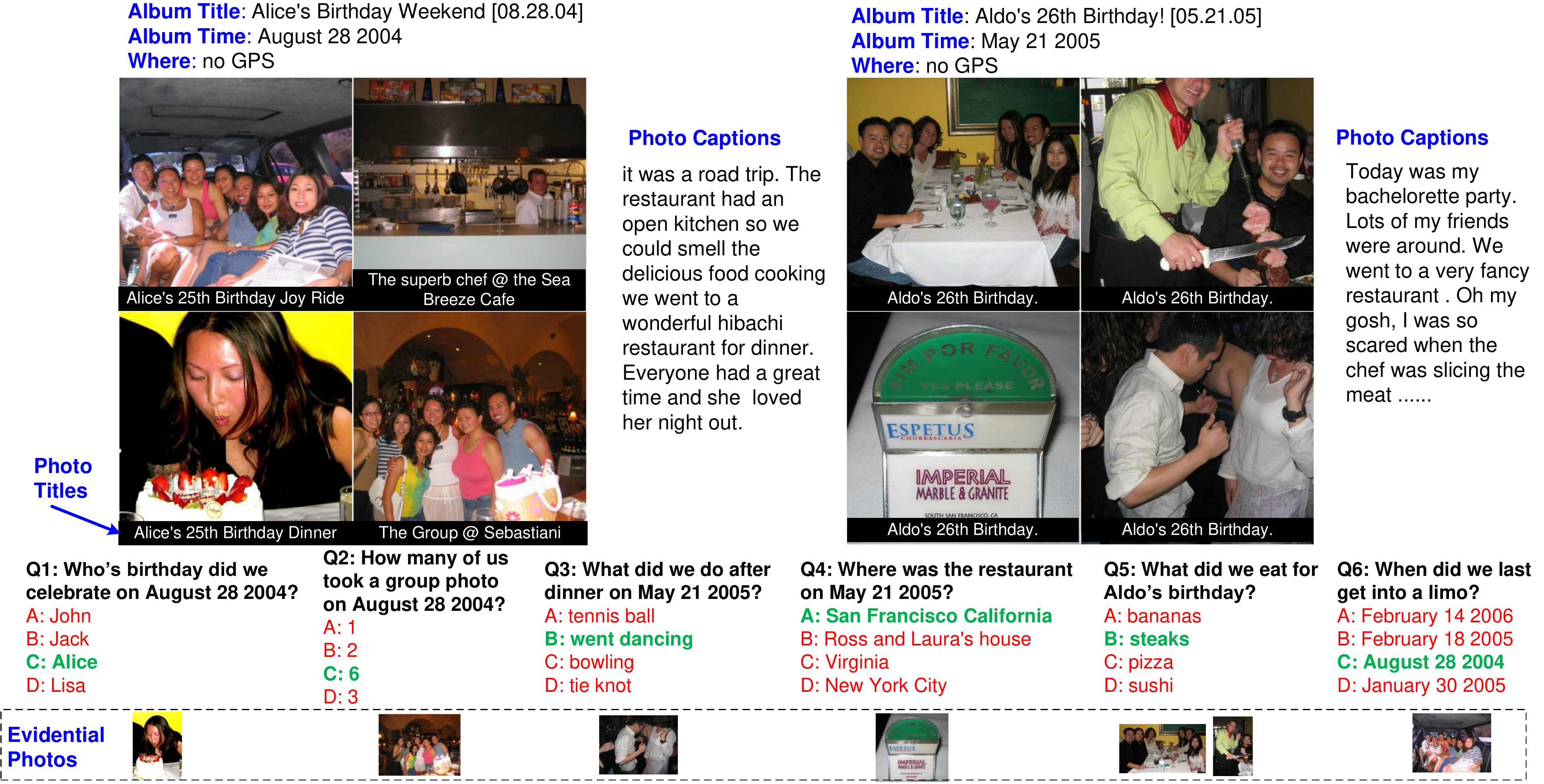}
\caption{\label{fig:qa_examples}Examples of QAs in the MemexQA dataset. From the top to bottom are album metadata, photos from two albums, photo titles and captions (right), question and candidate answers, evidential photos. The green answer is the ground-truth, and the red ones are incorrect candidate answers.
Q1-Q4 are questions about a single album, Q5-Q6 are about all albums.
}
\vspace{-6mm}
\end{figure}

This section introduces the MemexQA dataset
and compares it to existing VQA datasets.
The MemexQA dataset consists of 13,591 personal photos from 101 Flickr users. Theses personal photos capture a variety significant movements of their lives such as marriage proposals, wedding ceremonies, birthday/farewell parties, family reunions, commencements, etc. We collect comprehensive multimodal information about each photo, which includes a timestamp, GPS (if any), a photo title, photo tags, album information as well as photo captions from the SIND data~\cite{huang2016visual}. Through AMT (Amazon Mechanical Turk), we collect a total of 20,860 questions and 417,200 multiple choice answers. The annotators are instructed to first watch and understand all photos of a single Flickr user, treat these photos as if they were their own, and write meaningful questions\&answers that they might ask to recall their memories about the event in these photos. The annotator provides a correct answer as well as one or more evidential photos to justify the answer. See Fig.~\ref{fig:qa_examples}. In total 33,297 evidential photos are collected, i.e. 1.6 photos per question. The collected data are of decent quality, where the inter-human agreement is 91\%, which is comparable to 83.3\% on the VQA dataset~\cite{antol2015vqa}. 
Motivated by a recent study~\cite{jiang17delving}, we focus on five types of questions: ``what'', ``who'', ``where'', ``when'' and ``how many''. As the study showed, the query terms in the first 4W category account for more than 60\% of personal search traffic. We choose not to include the ``show me'' questions, as such questions can be addressed by the classical image/video retrieval method.

It is worth noting that MemexQA is significantly more expensive to collect than VQA. In VQA, annotators write a question by looking at only a single photo. In MemexQA, to write a question, they need to not only inspect tens of photos, but also consult the supporting multimodal information.
As a result, we estimate that it takes 10 times longer for annotators to write (96$\pm$10 seconds) and answer (62$\pm$3 seconds) a MemexQA question than a VQA question. Given those restrictions, we consider the MemexQA dataset to be sufficiently large as a first testbed on which different methods for this novel task can be compared. The experimental results in Section~\ref{sec:experiments} substantiate this claim.

\vspace{-2mm}
\subsection{Data Collection}
Generally, the MemexQA data are collected on AMT in 3 steps: QA collection, candidate answers generation, and QA cleanup. 
We acknowledge the following assumption for collecting the data: we assume all questions should be answerable by any individual and not just the owner of the photos. 
Therefore, information that is not captured by the photo or metadata should not be used in answering questions. This may lead to lower recall yet a more objective approach for evaluation.  

\noindent
\textbf{Questions \& Answers}. We select the Flickr users that have at least 4 albums from the SIND dataset~\cite{huang2016visual}. This results in a collection of 13,951 Flickr photos from 630 albums of 101 real Flickr users. 
The QA data collection tasks are conducted on AMT, an online crowd-sourcing platform. The online AMT annotators are instructed to write 1) important questions 2) objective questions, and 3) concise answers about the main topic in each album. Each annotator is only asked to write questions about a single Flickr user. Specifically,  they write 5 questions/answers about the photos in a single photo album, and for each answer they provide one or more evidential photos to justify their answers. This process is first repeated for all albums separately. Then the photos from all albums are shown, and they write questions about photos across multiple albums, and similarly provide evidential photos for each answer. The frequent questions about a single album are: ``where did we go'', ``what did we do'',  ``what was the wealth like'', etc. The frequent questions about multiple are ``what was the last time'', ``how many times did''.

\noindent
\textbf{Candidate Answers}. Following~\cite{zhu2016visual7w}, MemexQA employs both human workers and an automatic method to generate a pool of candidate answers. For the ``what'' question, a pool of candidate answers is automatically generated based on the answer of similar questions in the MemexQA and VQA datasets. For other types of questions, a pool of candidate answers is obtained by randomly selecting relevant user metadata. 
All candidate answers are then inspected by annotators to ensure there is only one correct answer for each question. 

\noindent
\textbf{QA Cleanup}
To reduce the ambiguous and low-quality QA, each photo album is independently annotated by at least 5 AMT workers. The QA along with candidate answers are verified by 3 workers, where they are asked to select the correct answer from the 4 choices. The workers report unreasonable QAs, when they find subjective questions/answers, or answers with 1+ correct choices. The collected data are of decent quality. The sampled inter-human agreement, which measures the percentage of the questions having same answer cross different AMT workers, is 0.9. This number is comparable or even better than that on existing VQA datasets~\cite{antol2015vqa,zhu2016visual7w}.

\vspace{-1mm}
\subsection{Comparison to VQA datasets}
\label{sec:dataset_comp}
\vspace{-1mm}

MemexQA is a real-world QA dataset that requires multimodal reasoning based on both vision and language. It compliments existing VQA datasets by allowing users to ask multimodal questions over a collection of photos. Table~\ref{tab:dataset_comp} compares MemexQA to representative VQA datasets~\cite{antol2015vqa,zhu2016visual7w}. 
As we see, the MemexQA dataset encompasses the most comprehensive information including evidential photos, Time GPS, and metadata. More importantly, it is the only effort that has questions grounded on multiple images. It further has the longest question and answers, which, according to~\cite{zhu2016visual7w}, is a good indicator for high-quality data.


MemexQA is a realistic dataset. It comprises personal photos of 101 real users and 20k faithful questions that users asked to recall their memories. We found that the statistics about MemexQA are closer to the statistics about Flickr photos, from which VQA and Visual7W datasets are crawled. For example, FlickrOR in Table~\ref{tab:dataset_comp} calculates the overlap ratio between the frequent search terms in the Flickr search logs~\cite{jiang17delving} and the frequent answers in QA datasets. As we see, MemexQA's ratio is 6\% higher than others, which suggests its answers are closer to 
user interests.
Similarly, the question category distribution of MemexQA also seems more realistic (see supplementary materials).


MemexQA is a multimodal dataset. The comprehensive multimodal information results in a ideal testbed for experimenting the vision+language QA problem. Fig.~\ref{fig:qa_examples} illustrates the diverse sources from which an answer can be derived, e.g. from title+time (Q1), vision+time (Q2, Q3, Q6), vision+OCR+GPS (Q4), vision+time+album info (Q5). 
As we show in Section~\ref{sec:experiments}, a solution to MemexQA may also be applicable to generic QA problems including text, image or video QA. 

\begin{table*}[ht]
\vspace{-3mm}
\centering
\caption{Comparison with the representative VQA and Visual7W dataset.}
\footnotesize
\label{tab:dataset_comp}
\begin{tabular}{c||ccccccc}
\hline
Dataset & \#Img & AvgQLen & AvgALen & FlickrOR& CrossImgs & Evidence& GPS/Meta\\ 
\hline
\hline
VQA~\cite{antol2015vqa}      & 204,721 & 6.2$\pm$2.0 & 1.1$\pm$0.4  & 39\%    & one   &  &\\
Visual7W~\cite{zhu2016visual7w}& 47,300  & 6.9$\pm$2.4 & 2.0$\pm$1.4  & 34\%    & one   & &\\
MemexQA  & 13,591  & 9.3$\pm$1.0 & 2.1$\pm$0.5  & 45\%  & many   &\checkmark &\checkmark\\
\hline
\end{tabular}
\vspace{-3mm}
\end{table*}

\vspace{-1mm}
\section{MemexNet} 
\vspace{-2mm}

\begin{figure*}[ht]
\vspace{-3mm}
\centering
\vspace{-3mm}\includegraphics[width=0.95\linewidth, height = 25mm]{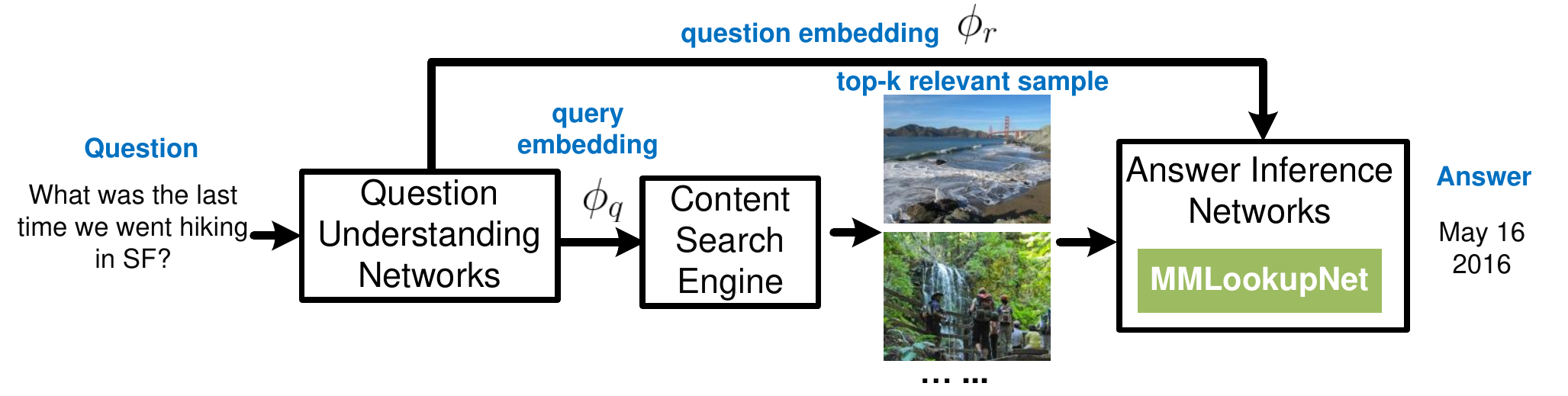}
\vspace{-3mm}
\caption{\label{fig:framework}Framework of MemexNet.}
\vspace{-5mm}
\end{figure*}

\label{sec:model}
This section introduces MemexNet, a novel network architecture inspired by the Watson QA system~\cite{ferrucci2010building}. MemexNet is designed for MemexQA, but may also be applied to text or video based question answering. As illustrated in Fig.~\ref{fig:framework}, there are three major modules in a MemexNet: the question understanding, search engine, and answer inference module. 

\noindent\textbf{Question understanding.} 
The query understanding module is responsible for encoding the question for subsequent search and inference tasks. Specifically, it contains two subnetworks, a \textit{question encoder} $\phi_r$ and a \textit{query encoder} $\phi_q$. Given a question $Q$, the question encoder embeds it into a latent state vector $\phi_r(Q) \in \mathbb{R}^{m_r}$, which expresses the question content and the estimated category to which its answer belongs, such as a date, number, name, place or action, scene, etc. $\phi_r(Q)$ will be passed to the answer inference module by a residual connection (shown in Fig.~\ref{fig:framework}). The query encoder discovers underlying concepts in the given question and produces a query embedding $\phi_q(Q)$ as the input to the content search engine. Usually, $\phi_q(Q)$ is a sparse vector over a set of relevant concepts in the vocabulary. For the multimedia cases, the query encoder maps a natural language sentence to a set of relevant concepts that exist in the pre-trained concept vocabulary $\mathbb{V}$~\cite{jiang17delving}. For pure textual retrieval, \textit{e.g.} for TextQA, the query encoder can be replaced by a simple lemmatization model, since out-of-vocabulary words are rare. For MemexNet we employ LSTM networks as our question encoder, SkipGram models~\cite{mikolov2013distributed} and VQE (Visual Query Embedding) networks~\cite{jiang17delving} as the query encoder. The supplementary materials discuss implementation details. 

\noindent\textbf{Search engine.} 
The second module is a content-based search engine that can automatically index and search the content of images/videos. Given a collection of samples $\mathbf{X} = \{\mathbf{x}_i\}$ (images or videos), it outputs a list of samples ranked by $P(\mathbf{x} | \phi_q(Q), \Theta)$, where $\phi_q$ is the embedded query and $\Theta$ represents the retrieval model, which retrieves images/videos based on the embedded query $\phi_q(Q)$. Instead of learning everything from scratch, we propose to leverage the off-the-shelf retrieval models for training the MemexNet. This is mainly to reduce the learning space for answer inference. The state-of-the-art visual search engines~\cite{sivic2003video,jiang2015bridging} allow us not only to search from large scale image or video corpus efficiently, but also to perform complex queries which include the time, geolocation, or subtle relation between concepts. In this work, we choose not to fine-tune the retrieval model but, instead, to learn attention weights over the retrieved samples. Indeed, an alternative approach is to learn a $k$-nearest neighbor function from scratch, as suggested in memory networks~\cite{chandar2016hierarchical}. However, we found that it might not improve the accuracy, as it might be too difficult to learn a neighborhood function that maps from text to images/videos in the current datasets.

\begin{figure*}[t]
\centering
\includegraphics[width=0.9\linewidth, height = 40mm]{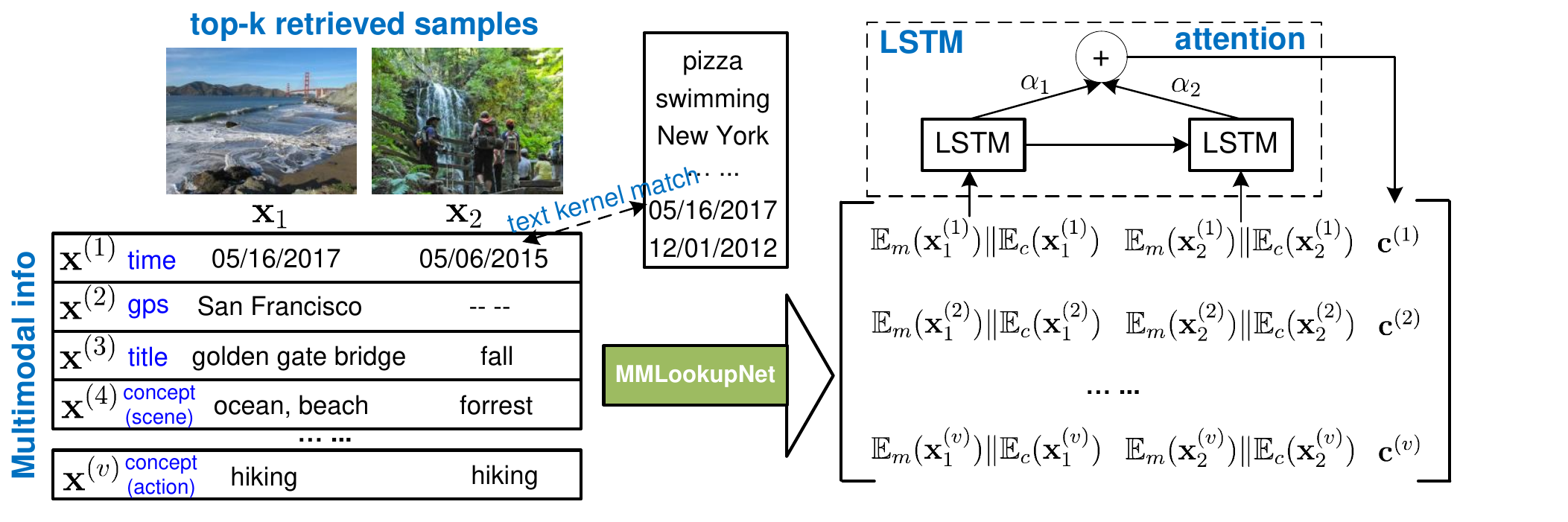}
\vspace{-3mm}
\caption{\label{fig:mmlookupnet}Illustration of the proposed MMLookupNet in the answer inference module.}
\vspace{-4mm}
\end{figure*}

\noindent\textbf{Answer inference.} 
The answer inference module is a modular network~\cite{andreas2016neural}, in which various networks can be plugged in/out for different QA tasks. For example, the Bi-Directional Attention Flow (BiDAF) layers are used for Text QA in Section~\ref{sec:exp_textqa}. To tackle the answer inference problem in MemexQA, we develop a novel MMLookupNet (MultiModal Lookup Network), shown in Fig.~\ref{fig:mmlookupnet}. Following VQA, in this paper, we regard MemexQA as a classification problem. For each question $Q$, we have the correct answer coded in the one-hot vector representation $\mathbf{y} \in \mathbb{R}^{m}$ of $m$ answer classes. We propose to estimate the probability of an answer class using the top-$k$ retrieved samples b y$\phi_r(Q)$:
\begin{equation}
\label{eq:prob_prediction}
P(\mathbf{y}| Q, \mathbf{X}) \!=\! P(\mathbf{y}| \phi_r(Q), \mathbf{x}_1,\! \cdots\!, \! \mathbf{x}_k) \!=\! f(\phi_r(Q), \Psi_{\text{lookup}}(\mathbf{x}_1, \! \cdots \!, \mathbf{x}_k), \text{conv}(\mathbf{x}_1), \! \cdots \!, \text{conv}(\mathbf{x}_k)),
\end{equation}
where $\mathbf{x}_i$, $i=1,\cdots,k$ are the top-$k$ samples ranked by $P(\mathbf{x}_i | \phi_q(Q), \Theta)$; $\Psi_{\text{lookup}}$ represents the outputs of MMLookupNet, and $\text{conv}(\mathbf{x}_i)$ is the CNN-based image feature for the sample $\mathbf{x}_i$. $f$ in Eq.~\eqref{eq:prob_prediction} represents the classification function, which, in MemexNet is implemented as two fully-connected layers.

For training, as shown in Eq.~\eqref{eq:prob_prediction}, 
the lookup embedding $\Psi_{\text{lookup}}(x_1,\cdots,x_k)$, the question embedding $\phi_r(Q)$ and the image CNN features are all fed to the classification layers $f$ to minimize the softmax cross-entropy loss, in which only the losses to the class in the multiple choices are considered.

 
The rest of the section will discuss the lookup embedding $\Psi_{\text{lookup}}(x_1,\cdots,x_k)$. The retrieved samples come with various metadata such as the timestamp, title, concepts, etc., as shown in Fig.~\ref{fig:mmlookupnet}. Let the superscript $\mathbf{x}^{(j)}$ index the $j$-th modality of $\mathbf{x}$. For example, $\mathbf{x}_1^{(1)}$ is ``05/16/2017'' in Fig.~\ref{fig:mmlookupnet}). MMLookupNet represents the multimodal information of the retrieved samples by concatenating two embeddings: $\mathbb{E}_{m}$ for modalities and $\mathbb{E}_{c}$ for matched classes. Specifically, same modalities share the same $\mathbb{E}_{m}(\mathbf{x}^{(j)}) \in \mathbb{R}^{m_{em}}$ indexed by the modality index $j$. $\mathbb{E}_{c}(\mathbf{x}^{(t)}) \in \mathbb{R}^{m_{ec}}$ is calculated from:
\begin{equation}
\label{eq:class_embedding}
\mathbb{E}_{c}(\mathbf{x}^{(j)}) = \mathbb{E}_{c}({\arg\max}_{i=1 \cdots m} \kappa(\mathbf{x}^{(j)}, a_i)),
\end{equation}
where $a_i$ is the answer text from the $i$-th class and $\kappa$ is a 
kernel function measuring the similarity between the text of $\mathbf{x}^{(j)}$ and $a_i$.
For this paper, we simply compute $\kappa$ as the proportion of exactly matched non-stop stemmed words. We introduce 
a pilot answer named ``\$'' where $\forall j \in [1,m], \kappa(\mathbf{x}^{(j)}, \$) = 0.5$. The pilot answer can gate non-match answers, \emph{i.e.} the cases where the best matched answer still has a very low $\kappa$ score.

Let $i \in [1,k]$ index the top-$k$ retrieved samples and $j \in [1,v]$ index the modality. Denote $\mathbf{e}_i^{(j)} = \lfloor \mathbb{E}_{m}(\mathbf{x}_i^{(j)}) \| \mathbb{E}_{c}(\mathbf{x}_i^{(j)}) \rfloor$ for the embedding about the $j$-th modality of the $i$th sample, where $\|$ is the vector concatenation operator. Intuitively, samples at different rank positions should have different importance. To incorporate this information, at the answer inference step, we introduce attentions over the retrieved samples. As shown in Fig.~\ref{fig:mmlookupnet}, $\mathbf{e}_i^{(j)}$ are sequentially fed into the LSTM network, where the attention weight $\alpha$ is calculated from:  
\begin{equation}
\label{eq:lookup_attention}
\mathbf{u}_{i} = \mathbf{v}^T \text{tanh}(W_1 \mathbf{h}_i + W_2 \phi_r(Q)), \alpha_{t} = \text{softmax}(\mathbf{u}_t),
\end{equation}
where $\mathbf{h}_i$ is the LSTM output at the ranked position $i$, and $\mathbf{v}, W_1,W_2$ are parameters to learn. The context vector $\mathbf{c}_k = \sum_{i=1}^k \alpha_{i} \mathbf{h}_i$ is used to represent the LSTM outputs with attention over rank positions. 
Similarly, we will use $j$ to index the context vector for different modalities. 
Finally, MMLookupNet represents the modality $j$ by a concatenated vector of $\mathbb{E}_{m}(\mathbf{x}_i^{(j)})$, $\mathbb{E}_{c}(\mathbf{x}_i^{(j)})$ and $\mathbf{c}_k^{(j)}$, where $i = 1, \! \cdots,\! k$. Then the above vectors of all modalities are concatenated as $\Psi_{\text{lookup}}$ to represent the top-$k$ sample. See Fig.~\ref{fig:mmlookupnet}.

$\Psi_{\text{lookup}}$ introduces an embedding space for models to learn to either lookup or infer an answer based on available information. When multiple matched answers are found, the model is expected to learn salient weights to attend to samples at appropriate ranked positions by $\alpha$, and meanwhile to differentiate modalities by the modality embedding $\mathbb{E}_{m}$. For example, the embedding $\mathbb{E}_{c}$ for the ``time'' modality is expected to receive higher weights for ``when'' questions. When no matched answers are found, the model still can infer an answer based on other information such as the question embedding, image CNN features and the modality embedding.

\vspace{-3mm}
\section{Experiments}
\vspace{-2mm}

\label{sec:experiments}
In this section, we empirically verify MemexNet on three tasks: MemexQA, TextQA and VideoQA. 

\subsection{MemexNet on MemexQA}\label{sec:exp_memexqa}

\textbf{Human Experiments} We first examine the human performance on MemexQA. We are interested in measuring 1) how well human can perform in the MemexQA task, 2) what is the contribution of each modality in helping users answer questions, and 3) how long does it take for humans to answer a MemexQA question. We conduct a set of controlled experiments, where AMT workers are asked to select an answer from 4 choices given different information, which includes \textit{Q}uestions, \textit{A}nswers, \textit{I}mages with timestamps and gps (if any), and \textit{M}etadata (titles and descriptions). Table~\ref{tab:exp_overall_memexqa} reports the results. As we see, humans manage to correctly guess 50\% of the correct answers using common sense. With all information, the accuracy reaches 0.93, which is comparable to 0.83 on VQA~\cite{antol2015vqa} and 0.97 on Visual7W~\cite{zhu2016visual7w}. The accuracy without images drops significantly, suggesting MemexQA requires substantial visual understanding and  reasoning. Besides, experiments show that on average it takes 10 times longer (62s) for humans to answer a MemexQA question than a VQA question (5.5s~\cite{zhu2016visual7w}). This suggests that the automatic MemexQA model, which can be consulted with exceeding speed and flexibility, can help provide enlarged supplement to one's memory.

\textbf{Model Experiments} We then study the  performance of state-of-the-art models on the MemexQA task\footnote{The experiments were performed on the MemexQA dataset v0.8 version.}. By default, all models use photos, timestamps, gps, and titles. Models are learned on a training set of 14,156 randomly selected QA pairs, and tested on a set of 3,539 QAs. The remaining QAs are used for validation. The most frequent 7,236 answers are selected as class labels, which cover 99\% of the multiple choices in the dataset.
To represent an image, we use the output of the last $fc$ layer of ResNet~\cite{he2016deep}, PCA-reduced to 300 dimensions (dimensionality reduction speeds up convergence). 

We compare MemexNet with both classical and state-of-the-art VQA models: \textit{Logistic Regression} predicts the answer from a concatenation of image, question and metadata features. Question and metadata features are represented by the averaged word embedding from the pre-trained SkipGram model~\cite{mikolov2013distributed}. \textit{Embedding} is similar to Logistic Regression except it learns the word embedding for questions and metedata. \textit{BoW} derives the answer by a concatenation of the image feature and the bag-of-words representation of the question and metadata. \textit{LSTM} represents the LSTM VQA model in~\cite{malinowski2015ask}, where the concatenation of question embedding, average image features, and average metadata embeddings are fed into every LSTM unit. \textit{LSTM + Att} represents the previous LSTM model with attention mechanisms. \textit{LSTM w./ Multi-Channels} is the RNN network described in~\cite{antol2015vqa}, where the sequence of question words, metadata, and photos are sequentially fed into three LSTM networks, respectively. The element-wise dot product of all hidden states are used for classification. For MemexNet, Eq.~\eqref{eq:prob_prediction} is used, where top-$k$ is set to 2. We extract concepts by the Google Vision API ({\small \url{https://cloud.google.com/vision/}}), employ E-Lamp Lite~\cite{jiang2015bridging} as the content search engine. and the SkipGram model as the query encoder. See the details in supplementary materials.  

Table~\ref{tab:exp_overall_memexqa} reports the overall performance. As we see, MemexNet outperforms all baseline methods, with statistically significant differences. The result indicates MemexNet  is a promising network for vision and language reasoning on this new task. The significant gap between the human and model accuracy, however, indicates that MemexQA is still a very challenging AI task. To analyze the contribution of the MMLookupNet, we replace it with the average embedding of concepts and metadata, and report the accuracy in Table~\ref{tab:exp_overall_memexqa}. As we see, the notable performance drop, especially on ``what'' and ``when'' questions, suggests MMLookupNet is beneficial in training MemexNet.

\begin{table*}[ht]
\vspace{-3mm}
\centering
\footnotesize
\caption{\label{tab:exp_overall_memexqa} Performance comparison on the MemexQA dataset}
\begin{tabular}{l||cccccc}
\hline
Method& how many & what & when & where & who&overall \\
\hline
\hline
Human (Q+A)&0.573&0.407&0.503&0.521&0.464&0.518\\
Human (Q+A+I)&0.928&0.727&0.901&0.846&0.755&0.863\\
Human (Q+A+M)&0.708&0.598&0.765&0.637&0.561&0.674\\
Human (Q+A+I+M)&0.941&0.865&0.961&0.961&0.860&0.927\\
\hline
\hline
BoW&0.698&0.217&0.222&0.269&0.252&0.290\\
Logistic Regression&0.645&0.241&0.217&0.277&0.26&0.295\\
Embedding&\textbf{0.719}&0.270&0.261&0.351&0.325&0.345\\
LSTM&0.758&0.340&0.336&0.332&0.348&0.390\\
LSTM + Att&0.700&0.356&0.461&0.391&0.451&0.433\\
LSTM w/. Multi-Channels&0.717&0.282&0.347&0.300&0.342&0.356\\
MemexNet w/o. MMLookupNet&0.677&0.378&0.382&0.353&0.430&0.418\\
MemexNet&0.668&\textbf{0.448}&\textbf{0.480}&\textbf{0.460}&\textbf{0.468}&\textbf{0.484}\\
\hline
\end{tabular}
\vspace{-3mm}
\end{table*}

\subsection{MemexNet on TextQA}\label{sec:exp_textqa}
This subsection evaluates MemexNet on TextQA on the SQuAD dataset~\cite{rajpurkar2016squad}. We employ a lemmatization model as our query encoder and the bi-directional LSTM networks (word embedding + char CNN) as our question encoder. In this data, we treat each sentence as a photo, each paragraph as an album, and retrieve the top 5 sentences relevant for each question using the language retrieval model with the JM smoother~\cite{zhai2004study}. The last 3 layers in the BiDAF model~\cite{seo2016bidirectional} are used as our answer inference networks. We compare MemexNet with recent TextQA networks in terms of 
$F_1$ score in Table~\ref{tab:squad_performance}. As we see, although MemexNet is not designed for this task and only uses 5 sentences, it achieves accuracy which is comparable or better than the recent text QA models. 

\begin{table*}[ht]
\vspace{-3mm}
\centering
\footnotesize
\caption{Performance comparison on the SQuAD development set}
\label{tab:squad_performance}
\begin{tabular}{c||cccccc}
\hline
Method   & LR    & Match-LSTM~\cite{wang2016machine} & DCN~\cite{xiong2016dynamicb}  & MPCM~\cite{wang2016multi}  & BiDAF~\cite{seo2016bidirectional}  & MemexNet\\
\hline
$F_1$ & 0.510 & 0.739  & 0.756 & 0.758 & 0.760 & \textbf{0.767}  \\    
\hline
\end{tabular}
\vspace{-3mm}
\end{table*}

\subsection{MemexNet on VideoQA}\label{sec:exp_videoqa}
This subsection verifies the efficacy and scalability on the VideoQA task. To this end, we apply MemexNet on YFCC100M, one of the largest public personal image and video collections~\cite{thomee2016yfcc100m}, and ask questions about 800K videos in the collection. We choose not to use the user-generated metadata (titles or descriptions) to access the QA performance based only on video content understanding. We build our VideoQA model using similar configurations as in Section~\ref{sec:exp_memexqa}. 
For such a large scale problem, an important metric is the 
runtime speed, which reflects the scalability and efficiency 
in real applications. To measure the run timespeed, we use a single core of Intel Xeon 2.53GHz CPU with 32GB memory. On average, 
it takes about 1.3 seconds to answer a question over 800 thousand videos purely based on the video content, suggesting that the proposed MemexNet is scalable for real-world large-scale QA problems. See some example questions in our demo video in the supplementary materials. Since the annotated ground truth is impossible to obtain for all the videos, we ask two human experts to ask our systems 25 questions and evaluate the results. The estimated averaged accuracy is 0.52. The breakdown of results is included in the supplementary material.



\section{Conclusions and Future Work}
\label{sec:conclusions}
In this paper, we propose MemexQA, a multi-disciplinary AI task and present a large dataset of real personal photos and crowd-sourced questions/answers to tackle that problem. We further present MemexNet, a unified, end-to-end trainable network architecture for image, text and video question answering. The experimental results on the MemexQA dataset demonstrate MemexNet's state-of-the-art performance on this challenging task. The promising results on TextQA and VideoQA suggest MemexNet is applicable to QA tasks across various domains and efficient for large scale image and video albums. We consider our work as a first step towards solving this new interdisciplinary AI problem. Our future works include personalized QA, fine-grained photo and album understanding. 

\section{Acknowledgments}
This work was partially supported by Yahoo InMind project and Flickr Computer Vision and Machine Learning group.

{\footnotesize
\bibliographystyle{abbrv}
\bibliography{egbib}
}

\end{document}


\title{Supplementary Materials of MemexQA: Visual Memex Question Answering}

\author{}

\maketitle

\section{4W Category Distribution}

\begin{figure}[ht]
\centering
\includegraphics[width=0.85\linewidth]{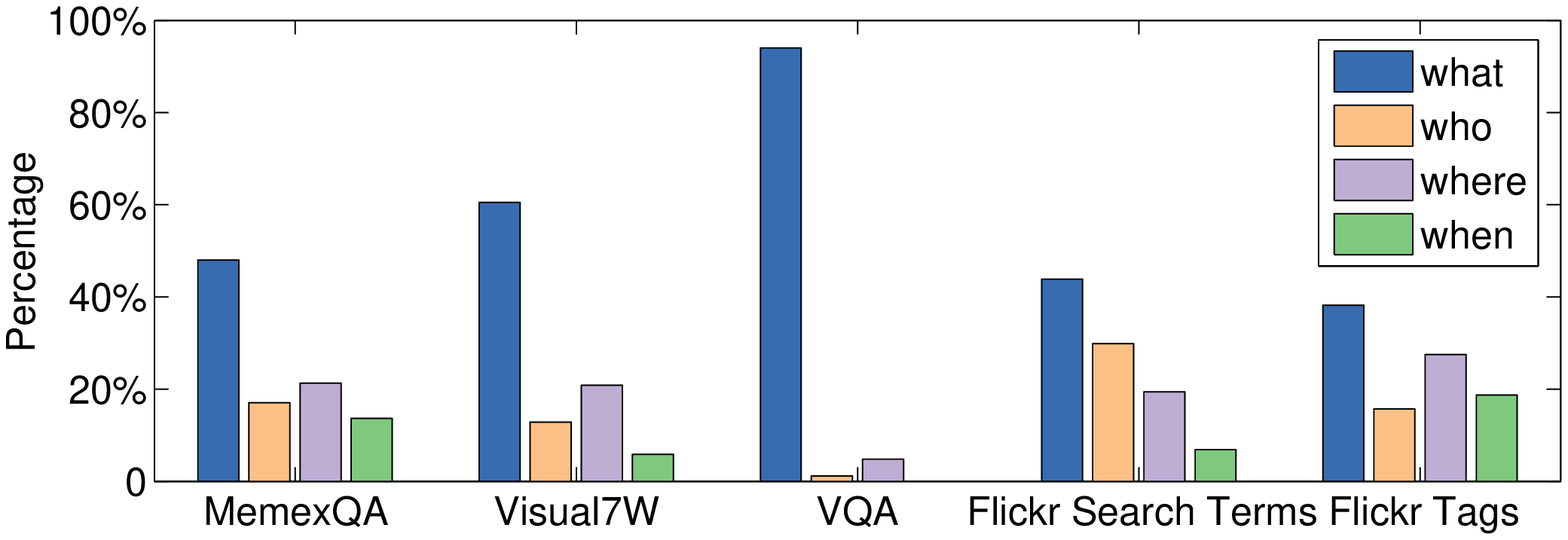}
\caption{\label{fig:4w}The 4W categories (what, when, who and where) distribution in each dataset.}
\end{figure}

We counted the answer in the 4W categories (what, when, who and where) in MemexQA, Visual7W and VQA datasets, and compared to the statistics with the 4W on the frequent search terms mined from Flickr search logs and Flickr user tags~\cite{jiang17delving}. Fig.~\ref{fig:4w} shows the results. We measured the distance in terms of the KL divergence, and found that the MemexQA's 4W distribution is the closest to both Flickr frequent search terms and Flickr tags. The result suggests MemexQA answers are closer to user interests on Flickr.

\section{Implementation Details}

We implemented all deep models in TensorFlow, which were trained over mini-batches of 32 samples. All models were trained on a training set of 14,156 QA pairs and tested on a test set of 3,539 QA pairs. The total number of class $m$ is 7,236. For a fair comparison, each model was trained for 45 epochs, i.e. about 20K iterations. The standard gradient decent algorithm was used to train the BoW, Logistic Regression and Embedding models. the adaptive sub-gradient (Adagrad) algorithm~\cite{duchi2011adaptive} was used to train other RNN models for faster convergence. The maximum length for the question was set to 12 and the maximum words/concepts for each modality was set to 8. Most questions asking about tens of photos, which could be too long for LSTM networks to encode. Since existing VQA models have not handled reasoning over multiple photos, in our baseline implementation, we randomly encoded 8 relevant photos for each question. The last fully-connected layer of ResNet~\cite{he2016deep}, reduced to 300-dimension by PCA, was used to represent an image. We found that both the original and reduced-dimension feature yield similar results but the dimension reduction speeds up the convergence. 

For the MemexNet, we first pre-trained our question encoder on the VQA dataset~\cite{antol2015vqa} to classify predefined answer types. See Table~\ref{tab:answer_type}. Then the question encoder was fine-tuned in the end-to-end MemexNet training as discussed in the paper. We use the pre-trained SkipGram model as the query encoder which maps the input sentence to 5 most relevant concepts. BM25 was used as our retrieval model for both visual concepts and textual metadata. We extracted concepts and OCR feature by Google Cloud Vision API. We found concepts features provide the biggest contribution whereas the OCR features are less useful. We implemented the attention mechanism based on~\cite{bahdanau2014neural}, the AttionWrapper in the TensorFlow contrib library.

The question embedding $\phi_r$ was carried out by the LSTM networks with the latent state size set to 100. The top-2 retrieved samples were considered. The embedding size of $\mathbb{E}_{c}$ and $\mathbb{E}_{m}$ were set to 10. We found the modality-specific embedding size might further improve the accuracy. The size of the attended hidden state $\mathbf{c}_k$ was set to 5. Then the CNN image features, $\phi_r$,  $\mathbb{E}_{m}$, $\mathbb{E}_{c}$ and $\mathbf{c}_k$ were concatenated for classification. While calculating $\mathbb{E}_{c}$, only the matched answers in the multiple choices were considered, since the class outside of the multiple choices would always have the 0 loss. Two fully connected layers $fc_1: \text{relu}(in-dim, 32)$ and $fc_2: \text{sigmoid}(32, 7236)$ were used for classification.

\begin{table}[ht]
\centering
\caption{Question and answer types in the proposed system.}
\label{tab:answer_type}
\footnotesize
\begin{tabular}{cc||l}
\hline
Question Type & Answer Type  & Example                 \\
\hline
\hline
when  & date, year, season, hour, etc. &  What was the last time we went hiking?\\\hline
where & scene, gps, city, country, etc. & Where was my brother's graduation ceremony in 2013? \\\hline
what & action, object, activity, etc. & What did we play during this spring break? \\\hline
who   & name, face, etc. & Who did I meet in NIPS 2016? \\\hline
how many & number & How many times have I had sushi last month? \\\hline
yes/no & yes, no & Did I do yoga yesterday? \\\hline
\end{tabular}
\end{table}

\section{Experimental Results on Video QA}
\begin{figure}[ht]
\centering
\includegraphics[width=0.85\linewidth]{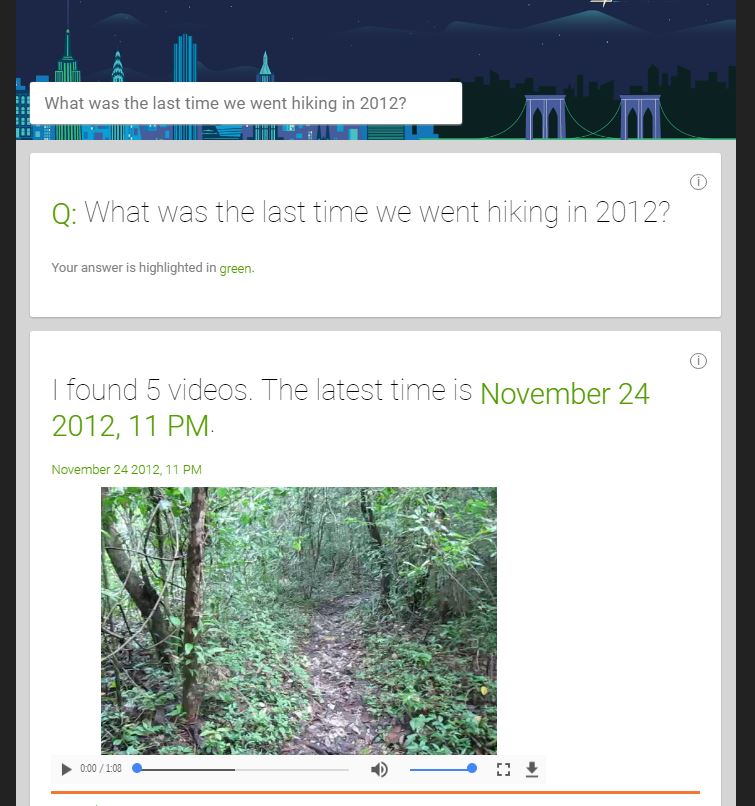}
\caption{\label{fig:videoqa_interface}The interface of MemexNet on the YFCC100M dataset}
\end{figure}

The goal of the experiment is to validate the efficacy and scalability, which is evaluated as the runtime second over the selected 25 questions. The interface of the prototype system can be found in Fig.~\ref{fig:videoqa_interface}. The experiments were conducted on a single core of Intel Xeon 2.53GHz CPU with 32GB memory. Table~\ref{tab:videoqa_performance} reports the runtime results. As we see, it takes about 1.3 seconds to answer a question over 800 thousand videos only based on the video content. The results indicate that the proposed MemexNet is scalable for real-world large-scale QA problems. Besides, to estimate the accuracy without the ground truth data, we ask 2 annotators to judge whether the returned answers are reasonable and report the accuracy in Table~\ref{tab:videoqa_performance}. \textbf{See the demo video in the supplementary material}.

\begin{table*}[ht]
\centering
\caption{The scalability and efficiency test of MemexNet on 800k videos.}
\label{tab:videoqa_performance}
\begin{tabular}{l||cccccc}
\hline
& How many & what& when& where& who& overall\\
\hline
Estimated accuracy& 1/5& 2/5  & 7/8  & 3/5 & 0/2 & 0.52 (13/25) \\
Runtime (s) & 2.21 & 1.91 & 0.87 & 1.15 & 0.79 & 1.39 \\
\hline
\end{tabular}
\end{table*}

{\footnotesize
\bibliographystyle{abbrv}
\bibliography{egbib}
}